\documentclass[11pt]{article}

\usepackage[preprint]{acl}

\usepackage{times}
\usepackage{latexsym}

\usepackage[T1]{fontenc}

\usepackage[utf8]{inputenc}

\usepackage{microtype}

\usepackage{inconsolata}

\usepackage{graphicx}
\usepackage{hyperref}
\usepackage{enumitem}
\usepackage{soul}
\usepackage{pifont}
\usepackage{xcolor} 
\usepackage{booktabs}
\usepackage{multirow}
\usepackage{fontawesome}
\usepackage[most]{tcolorbox}
\usepackage{color}      
\usepackage{fancyvrb}   

\definecolor{deepgreen}{rgb}{0.0, 0.6, 0.0}
\newcommand{\cmark}{\textcolor{deepgreen}{\ding{51}}}
\newcommand{\xmark}{\textcolor{red}{\ding{55}}}

\setlist[itemize]{topsep=0pt, partopsep=0pt, parsep=0pt, itemsep=0pt, leftmargin=10pt}

\title{Agent-Omni: Test-Time Multimodal Reasoning via Model Coordination\\for Understanding Anything}

\author{
\begin{tabular}{c}
Huawei Lin$^{1,2}$\hspace{.2in}
Yunzhi Shi$^{1}$\hspace{.2in}
Tong Geng$^{3}$\hspace{.2in}
Weijie Zhao$^{2}$\\
Wei Wang$^{1}$\hspace{.2in}
Ravender Pal Singh$^{1}$ \\
\end{tabular}\\
$^1$ Amazon \\
$^2$ Rochester Institute of Technology \\
$^3$ University of Rochester
}

\begin{document}
\maketitle

\begin{abstract}
Multimodal large language models (MLLMs) have shown strong capabilities but remain limited to fixed modality pairs and require costly fine-tuning with large aligned datasets. Building fully omni-capable models that can integrate text, images, audio, and video remains impractical and lacks robust reasoning support. In this paper, we propose an \textbf{Agent-Omni} framework that coordinates existing foundation models through a master-agent system, enabling flexible multimodal reasoning without retraining. The master agent interprets user intent, delegates subtasks to modality-specific agents, and integrates their outputs into coherent responses. Extensive experiments across text, image, audio, video, and omni benchmarks show that Agent-Omni consistently achieves state-of-the-art performance, particularly on tasks requiring complex cross-modal reasoning. Its agent-based design enables seamless integration of specialized foundation models, ensuring adaptability to diverse inputs while maintaining transparency and interpretability. In addition, the framework is modular and easily extensible, allowing future improvements as stronger models become available. 
\end{abstract}

\section{Introduction}

Multimodal large language models (MLLMs) extend the capabilities of language models by integrating text with other modalities, such as image~\cite{DBLP:conf/acl/ZhangY0L0C024, visionllama}, audio~\cite{kimi-audio, qwen2-audio}, and video~\cite{video-llava, llava-next, DBLP:conf/acl/XuGHAAFMZ21}. Existing systems are often restricted to fixed pairs, for example, text–image for captioning and visual question answering~\cite{DBLP:conf/cvpr/Guo0LT0TH23, llava}, text–video for event understanding~\cite{video-llava}, or text–audio for transcription and dialogue~\cite{kimi-audio, qwen2-audio}. However, in practice, many scenarios demand omni LLMs that can flexibly accept any combination of text, image, video, and audio while producing textual outputs~\cite{qwen2.5-omni, ming-omni}. For instance, a user might provide a background speech recording, an accompanying image, and a written note, then pose a question whose answer requires reasoning over all of these inputs~\cite{Nexus-O, Megrez-Omni}.

\begin{figure}[t]
\vspace{-.25in}
\hspace{-.2in}
    \includegraphics[width=1.1\linewidth]{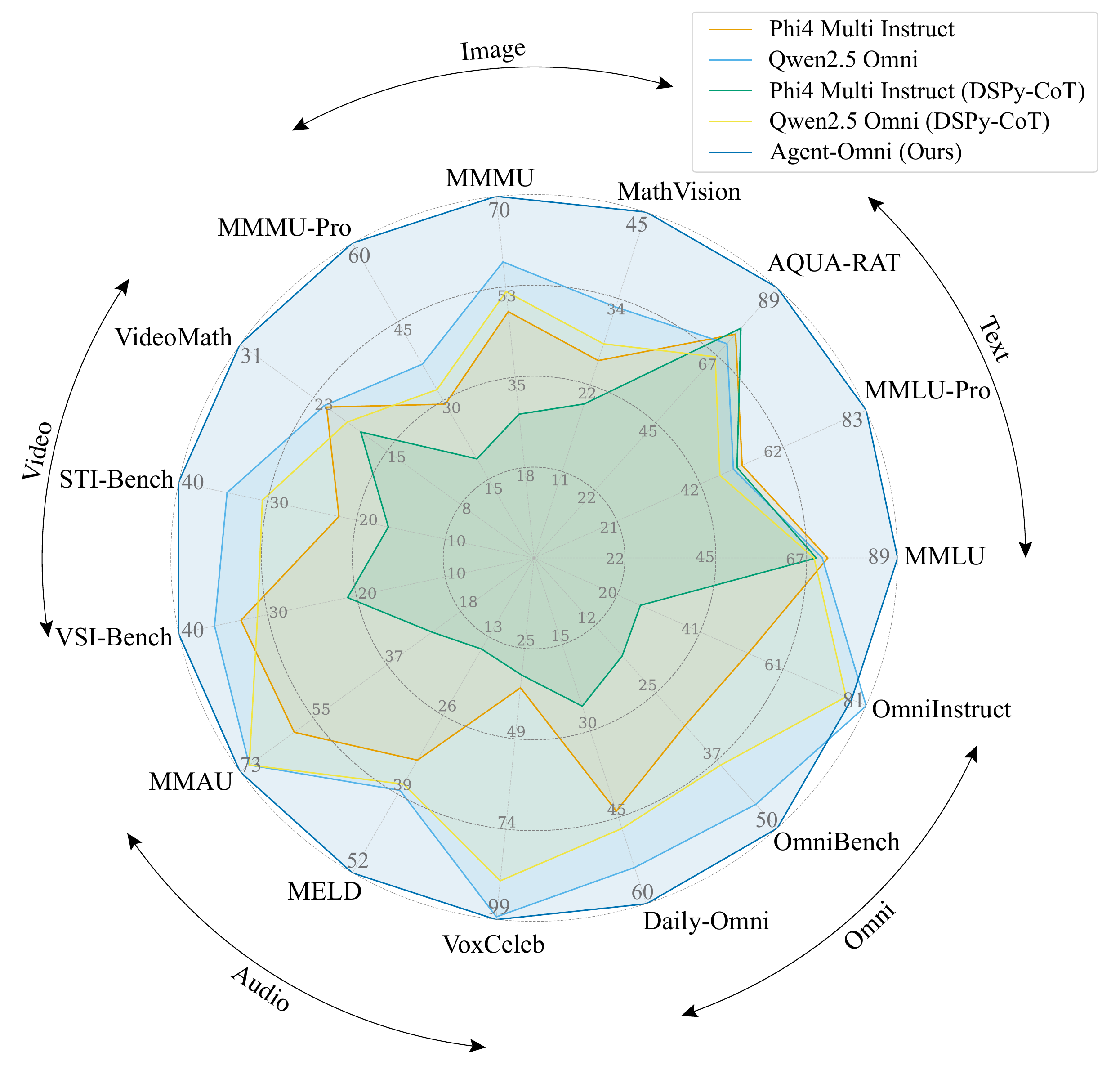}
    \vspace{-.3in}
    \caption{Comparison of Agent-Omni and other omni methods across multimodal benchmarks.}
    \label{fig:radar_chart}
    \vspace{-.2in}
\end{figure}

Extending existing multimodal LLMs into fully omni-capable systems typically requires large-scale fine-tuning across all modalities~\cite{llava-vl, video-llava, DBLP:journals/pami/ZengZLWZZ24, DBLP:journals/corr/abs-2412-08637}. This process demands extensive datasets that cover diverse cross-modal combination, and significant computational resources to jointly optimize model parameters. However, collecting omni-level training data that includes text, images, videos, and audio in aligned contexts is extremely costly and often impractical~\cite{qwen2.5-omni, ming-omni}. Moreover, even when such data is available and large-scale training is performed, omni models often suffer from trade-offs across modalities: improving performance on one modality can degrade accuracy on others, and balancing these objectives becomes increasingly difficult as the number of supported modalities grows~\cite{DBLP:journals/corr/abs-2309-10313, DBLP:journals/corr/abs-2505-19616}.

\begin{figure*}[t]
    \centering
    \includegraphics[width=\linewidth]{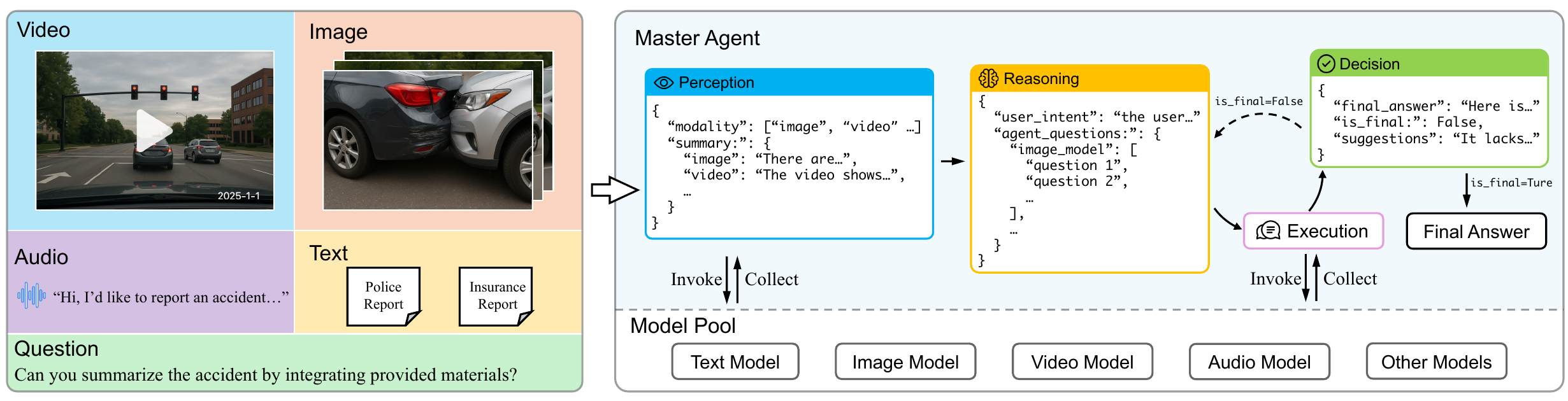}
    \caption{Overview of the Agent-Omni framework. A master agent interprets the query, identifies relevant modalities, and delegates sub-questions to corresponding foundation models (text, image, audio, video). Their outputs are iteratively integrated and refined through a self-improvement loop, enabling coherent multimodal reasoning in test-time inference.}
    \label{fig:overview}
    \vspace{-.15in}
\end{figure*}

Beyond training challenges, achieving effective omni reasoning is itself a difficult problem. In existing multimodal tasks such as visual question answering, and video understanding, reasoning has been shown to improve performance by enabling models to better connect information across paired modalities~\cite{DBLP:journals/tmlr/KeJMNXLLQWSXJ25, DBLP:journals/corr/abs-2412-09078}. Extending this ability to an omni setting is far more challenging: the system must integrate arbitrary combinations of modalities into a coherent understanding, for example aligning spoken descriptions with visual evidence or linking video events with accompanying text. Building datasets that support such reasoning is even harder than collecting aligned inputs, since they must include rich multi-source evidence and tasks requiring cross-modal integration. Because of this lack of datasets and the complexity of the problem, current approaches remain restricted to pairwise settings, and no truly omni reasoning model yet exists.  Table~\ref{tbl:head-table} summarizes the capabilities of representative models, highlighting the gap in achieving both broad multimodal coverage and strong reasoning ability.

\noindent\textbf{Limitations.} We identify the main limitations and challenges as follows:
\begin{itemize}
    \item \textbf{Heavy reliance on fine-tuning:} Building omni LLMs requires large-scale curated data across modalities and substantial computation, making training costly and impractical.
    \item  \textbf{Trade-offs between modalities:} Improving performance on one modality often leads to degradation in others, making optimization difficult.
    \item  \textbf{Lack of omni reasoning:} While reasoning improves performance in pairwise multimodal tasks, no models can reliably integrate arbitrary modality combinations. 
    \item  \textbf{Insufficient datasets:} Datasets supporting omni reasoning are largely unavailable, restricting current systems to limited modality pairs.
\end{itemize}

\begin{table}[t]
\caption{\label{tbl:head-table} Comparison of multimodal coverage and reasoning ability (\cmark: supported, \xmark: not supported)}
\vspace{-.1in}
\centering
\resizebox{.46\textwidth}{!}{%
\begin{tabular}{lcccc}
\toprule
\midrule
Method & Text   & \begin{tabular}[c]{@{}c@{}}Visual\\ (image \& Video)\end{tabular} & Audio  & Reasoning \\\midrule
Claude 3.7 Sonnet            & \cmark & \cmark                                                           & \xmark & \cmark    \\
Deepseek R1                  & \cmark & \xmark                                                           & \xmark & \cmark    \\
GPT OSS 20B                  & \cmark & \xmark                                                           & \xmark & \cmark    \\
Phi4 Multimodal Instruct    & \cmark & \cmark                                                           & \cmark & \xmark    \\
QWen2.5 Omni                 & \cmark & \cmark                                                           & \cmark & \xmark    \\
QWen2 Audio 7B	& \cmark& 	\xmark	& \cmark& 	\xmark\\
QWen3 4B Instruct            & \cmark & \xmark                                                           & \xmark & \xmark    \\
QWen3 4B Thinking            & \cmark & \xmark                                                           & \xmark & \cmark    \\
Llava Video 7B	& \cmark	& \cmark	& \xmark	& \xmark\\
Ours (Agent-Omni)            & \cmark & \cmark                                                           & \cmark & \cmark   \\
\midrule
\bottomrule
\end{tabular}
}%
\vspace{-.2in}
\end{table}

In this paper, we propose a omni agent that can ``understand anything'', i.e., interpret and answer user questions about any combination of text, image, audio, or video inputs by coordinating existing foundation models through dynamic agents, without fine-tuning or retraining.

\noindent\textbf{Contributions.} The main contributions of this paper are summarized as follows:
\begin{itemize}
\item   We propose a novel agent-based framework that coordinates existing foundation models to reason jointly over text, images, video, and audio, without any task-specific fine-tuning or retraining.
\item   We design a flexible master-agent system that interprets user intent, delegates subtasks to modality-specific agents, and integrates their outputs into a coherent final answer.
\item   We validate the framework through practical scenarios involving complex multimodal understanding and benchmark its performance across diverse tasks and datasets.
\item   We provide an open-source implementation\footnote{\mbox{\url{https://github.com/huawei-lin/Agent-Omni}}}, enabling future research and applications.
\end{itemize}

\section{Agent-Omni}
The goal of Agent-Omni is to enable flexible multimodal reasoning at test time without the need for retraining or large-scale fine-tuning. Instead of relying on a single unified model, our framework coordinates existing foundation models through a hierarchical agent architecture. By doing so, Agent-Omni can accept arbitrary combinations of text, images, audio, and video inputs, and produce coherent textual outputs.

\subsection{Overview}

Figure~\ref{fig:overview} illustrates the overall workflow. Consider the case where the user provides accident-related materials, including several photos, a dashcam video, an emergency call recording, and two documents (a police report and an insurance report), and asks: \textit{``Can you summarize the accident by integrating provided materials?''} The master agent identifies relevant modalities (image, video, audio, and text). It then formulates sub-questions for each modality, delegates them to the corresponding foundation models, and collects their structured outputs. These are iteratively fused through a self-improvement loop that resolves inconsistencies and refines the answer before final output.

\subsection{Master Agent}
The Master Agent serves as the central controller of Agent-Omni, with its internal workflow divided into four functional stages: 
(1) \textbf{Perception}: analyzes the input modalities and produces structured representations; 
(2) \textbf{Reasoning}: decomposes the user query into sub-questions based on the perceived information; 
(3) \textbf{Execution}: invokes appropriate foundation models from the model pool to answer the sub-questions and gathers their outputs; 
and (4) \textbf{Decision}: integrates all outputs to construct a final answer, evaluates its completeness, and determines whether another iteration of the reasoning loop is required.

\paragraph{Perception.}
The perception stage addresses the fundamental requirement that the agent must first understand what materials are provided before any reasoning can take place. 
As shown in Figure~\ref{fig:overview}, the Master Agent examines multimodal inputs (e.g., text, image, audio, video) and summarizes them into a concise \texttt{JSON} structure, where each modality is represented with a semantic description. 
This transformation turns raw signals into structured representations, ensuring that heterogeneous modalities are consistently aligned within a unified representation space. 
The generated \texttt{JSON} thus serves as the foundation for subsequent reasoning steps.

\paragraph{Reasoning.} 
After perception, the agent must decide how to utilize the perceived information to address the user's intent. In this stage, the Master Agent first derives a high-level \texttt{user\_intent} that summarizes what the user is asking.  It then formulates modality-specific sub-questions for each input modality that requires further reasoning. 
As illustrated in Figure~\ref{fig:overview}, these results are organized in a structured \texttt{JSON} format, where the user intent is explicitly represented, and the sub-questions are grouped under the corresponding modality (e.g., \texttt{image\_model}, \texttt{text\_model}). 
This design makes the reasoning process explicit and interpretable, while also providing a clear execution plan that connects each sub-question to its designated model. 

\paragraph{Execution.} 
Once the reasoning stage has produced modality-specific sub-questions, the execution stage faithfully carries them out by invoking the designated foundation models specified in the reasoning \texttt{JSON}. 
The outputs are then systematically collected and organized so that each sub-question is explicitly paired with its corresponding answer. 
This mechanism guarantees that intermediate results remain well-documented and easily traceable, thereby providing the factual grounding required for the subsequent decision stage.

\paragraph{Decision.} 
After execution has gathered answers from the foundation models, the agent must consolidate these results into a coherent response. 
In the decision stage, the Master Agent integrates all outputs recorded in the \texttt{JSON} structure to construct an answer for the user's original query. 
This answer is then evaluated for completeness and reliability: if gaps or inconsistencies are detected, the agent appends feedback instructions to the \texttt{JSON} and triggers another round of the reasoning–execution–decision loop (the master loop). 
As illustrated in Figure~\ref{fig:overview}, the decision stage produces three key components: (1) \texttt{final\_answer}, which provides a direct response to the original query; (2) \texttt{is\_final}, a flag indicating whether further iterations of the master loop are required -- if true, the \texttt{final\_answer} is returned as the final output; and (3) \texttt{suggestions}, which specify how subsequent iterations should refine the response if additional loops are necessary. 
This design enables iterative self-correction, ensuring that the final output is not only comprehensive but also progressively improved through repeated evaluation and refinement. 

\paragraph{Further Master Loop.} 
If the decision stage sets \texttt{is\_final} to \texttt{false}, the Master Agent initiates another iteration of the master loop. 
In the new loop, the reasoning stage consults the \texttt{suggestions} generated by the previous decision to prepare follow-up questions that target missing or uncertain information. 
These new sub-questions are then processed in the execution stage, whose outputs are passed again to the decision stage for integration and evaluation. 
This process repeats iteratively until either \texttt{is\_final} is set to \texttt{true}, indicating that the response is sufficiently complete, or a predefined maximum number of loops $L$ is reached. 
Such a design allows the Master Agent to progressively refine its answers through self-correction.

\begin{table*}[t]
\vspace{-.1in}
\caption{Accuracy on text benchmarks (MMLU, MMLU-Pro, AQUA-RAT).}\label{tbl:acc_text_modality}
\vspace{-.1in}
\resizebox{\textwidth}{!}{%
\begin{tabular}{ccccccccc}
\toprule
\midrule
Method                            & Model                     & \begin{tabular}[c]{@{}c@{}}MMLU\\ (STEM)\end{tabular} & \begin{tabular}[c]{@{}c@{}}MMLU\\ (Social Sciences)\end{tabular} & \begin{tabular}[c]{@{}c@{}}MMLU\\ (Humanities)\end{tabular} & \begin{tabular}[c]{@{}c@{}}MMLU\\ (Other)\end{tabular} & \begin{tabular}[c]{@{}c@{}}MMLU\\ (Average)\end{tabular} & MMLU-Pro         & AQUA-RAT         \\
\midrule
\multirow{7}{*}{\begin{tabular}[c]{@{}c@{}}Foundation\\ Model\end{tabular}} & Claude 3.7 Sonnet         & 91.87\%          & 90.49\%                & 81.89\%           & 87.88\%          & 88.03\%          & 76.75\%          & 87.40\%          \\
                                  & Deepseek R1               & \textbf{95.19\%} & \textbf{92.28\%}       & \ul{84.17\%}     & \textbf{91.25\%} & \textbf{90.72\%} & \ul{82.66\%}    & 87.40\%          \\
                                  & GPT OSS 20B               & 91.91\%          & 85.26\%                & 81.93\%           & 83.87\%          & 85.74\%          & 74.41\%          & 88.50\%          \\
                                  & Phi4 Multimodal Instruct & 75.77\%          & 75.56\%                & 65.19\%           & 71.80\%          & 72.08\%          & 52.13\%          & 74.02\%          \\
                                  & QWen2.5 Omni              & 74.43\%          & 74.34\%                & 62.33\%           & 71.83\%          & 70.73\%          & 49.93\%          & 70.87\%          \\
                                  & QWen3 4B Instruct         & 89.49\%          & 81.64\%                & 70.34\%           & 78.67\%          & 80.04\%          & 68.53\%          & 85.43\%          \\
                                  & QWen3 4B Thinking         & 91.05\%          & 82.91\%                & 72.29\%           & 81.04\%          & 81.82\%          & 67.64\%          & 86.61\%          \\
\midrule
\multirow{5}{*}{DSPy-CoT}         & Claude 3.7 Sonnet         & 92.17\%          & 90.44\%                & 82.51\%           & 89.76\%          & 88.72\%          & 78.48\%          & 84.65\%          \\
                                  & Deepseek R1               & 92.96\%          & \ul{91.81\%}          & \textbf{84.39\%}  & \ul{91.03\%}    & \ul{90.05\%}    & 74.83\%          & 88.58\%          \\
                                  & QWen2.5 Omni              & 72.87\%          & 72.08\%                & 59.80\%           & 70.13\%          & 68.72\%          & 46.54\%          & 66.54\%          \\
                                  & QWen3 4B Instruct         & 87.59\%          & 81.07\%                & 70.81\%           & 76.28\%          & 78.94\%          & 65.83\%          & 86.61\%          \\
                                  & QWen3 4B Thinking         & 92.51\%          & 83.66\%                & 75.37\%           & 81.73\%          & 83.32\%          & 69.85\%          & \textbf{89.76\%} \\
\midrule
\multicolumn{2}{c}{Ours (Agent-Omni)}                         & \ul{94.52\%}    & 90.40\%                & 81.68\%           & 90.31\%          & 89.23\%          & \textbf{83.21\%} & \ul{89.37\%}   \\
\midrule
\bottomrule
\end{tabular}%
}

\vspace{-.1in}
\end{table*}

\subsection{Model Pool} 
The model pool serves as the resource hub that provides the Master Agent with diverse foundation models to address modality-specific sub-questions. 
It contains a collection of specialized models spanning different input modalities, such as large language models for text, vision-language models for images, speech-text models for audio, and multimodal models capable of cross-modal reasoning. 
Each model in the pool can be invoked on demand during the execution stage, according to the reasoning plan specified in the \texttt{JSON}. 

Unlike conventional multimodal LLMs that require costly joint training or fine-tuning across all modalities, the model pool in Agent-Omni operates without any additional training. 
Existing foundation models are coordinated at inference time, making the framework both flexible and lightweight. 
New models can be seamlessly added to expand the agent’s capabilities, while existing ones are selectively leveraged based on their strengths. 
By decoupling model selection from reasoning, the Master Agent can dynamically orchestrate heterogeneous models in a unified workflow. 
This training-free design enables Agent-Omni to adapt to a wide range of multimodal queries while maintaining transparency, scalability, and efficiency.

\section{Experimental Evaluation}

The goal of our experiments is to systematically assess the performance of Agent-Omni across multiple dimensions of multimodal reasoning. We focus on the following four research questions: 
\textbf{(1)} How well does Agent-Omni generalize across diverse modalities (text, image, audio, and video), and can it achieve competitive performance on omni-level tasks? \textbf{(2)} What is the computational cost of the proposed Agent-Omni at test time, and how efficient is the framework compared to end-to-end multimodal LLMs? 
\textbf{(3)} How does the choice of different foundation models in the model pool affect the accuracy of Agent-Omni? and 
\textbf{(4)} How does varying the maximum number of master loops influence final performance, and to what extent does iterative self-correction improve answer quality?

\subsection{Datasets}
To comprehensively evaluate the multimodal understanding capability of Agent-Omni, we experiment on a broad collection of benchmarks that span five major categories:
\textbf{(1) Text.} We adopt classic language understanding and reasoning benchmarks, including MMLU~\cite{mmlu} (covering STEM, Social Sciences, Humanities, and Other domains), its more challenging variant MMLU-Pro~\cite{mmlu-pro}, and the arithmetic reasoning dataset AQUA-RAT~\cite{aqua-rat}.
\textbf{(2) Image.} For visual reasoning, we evaluate on MathVision~\cite{mathvision}, a benchmark targeting mathematical understanding from images, as well as MMMU~\cite{mmmu} and its robust extension MMMU-Pro~\cite{mmmu-pro}, which focus on expert-level multimodal and multidisciplinary reasoning.
\textbf{(3) Video.} To assess temporal and spatial reasoning, we include VideoMathQA~\cite{VideoMathQA}, which benchmarks mathematical problem-solving from videos, STI-Bench~\cite{STI-Bench}, designed for precise spatio-temporal understanding, and VSI-Bench~\cite{VSI-Bench}, which emphasizes video-based scene interpretation.
\textbf{(4) Audio.} For auditory understanding, we use MMAU~\cite{mmau}, a large-scale multi-task audio reasoning benchmark, MELD-Emotion~\cite{meld}, which evaluates emotion recognition in conversations, and VoxCeleb-Gender~\cite{Voxceleb}, a dataset for speaker gender classification.
\textbf{(5) Omni-level.} Finally, to test holistic multimodal reasoning that integrates multiple modalities, we include Daily-Omni~\cite{Daily-Omni}, which emphasizes cross-modal temporal alignment, OmniBench~\cite{OmniBench}, targeting universal multimodal understanding, and OmniInstruct~\cite{OmniBench}, a large-scale instruction-following dataset across modalities.

\subsection{Baselines}
To evaluate the performance of Agent-Omni, we compare it against two categories of baselines: \textbf{(1) Foundation Models:} We directly evaluate a set of state-of-the-art foundation models across modalities, including large language models, vision-language models, and other modality-specific models. 
\textbf{(2) DSPy-CoT:}  We adopt DSPy with chain-of-thought prompting as a strong baseline~\cite{dspy}. DSPy-CoT represents a method of improving reasoning within a single model by leveraging structured prompts, without introducing cross-model orchestration. This baseline highlights the difference between enhancing reasoning inside one model versus coordinating multiple specialized models, enabling a fair comparison of Agent-Omni.

\subsection{Models}
We evaluate Agent-Omni using a diverse set of state-of-the-art foundation models, each specialized in different modalities. 
\textbf{(1) Text:} large language models including Deepseek R1~\cite{deepseek-r1}, GPT OSS 20B~\cite{gptoss}, and QWen3 4B~\cite{qwen3}, which provide strong reasoning and problem-solving capabilities on text-centric benchmarks.  
\textbf{(2) Image \& Video:} vision-language models such as Claude 3.7 Sonnet and Llava Video 7B~\cite{video-llava}, designed to align visual and textual representations for tasks like image question answering and video understanding. 
\textbf{(3) Audio:} audio-focused models such as Qwen2 Audio 7B~\cite{qwen2-audio}, specialized in speech recognition and auditory reasoning.  
\textbf{(4) Omni:} multimodal models including Phi4 Multimodal Instruct~\cite{phi4} and Qwen2.5 Omni~\cite{qwen2.5-omni}, which natively support multiple modalities but often face trade-offs in robustness across them.  

As summarized in Table~\ref{tbl:head-table}, each model demonstrates strengths in its target modality but lacks full coverage across text, visual, audio, and reasoning dimensions. This highlights the motivation for Agent-Omni, which coordinates these specialized models to achieve balanced omni-modal reasoning.

\begin{table}[t]
\centering
\caption{Setup of Agent-Omni with selected foundation models and their roles.}
\vspace{-.1in}
\resizebox{.5\textwidth}{!}{%
\begin{tabular}{ccc}
\toprule
\midrule
Modality & Model             & Role / Function \\
\midrule
Master   & Claude 3.7 Sonnet & Central controller for reasoning and decision-making \\
Text     & Deepseek R1       & Strong LLM for text understanding and logical reasoning \\
Image    & Claude 3.7 Sonnet & Handles visual perception and image-based reasoning \\
Video    & Claude 3.7 Sonnet & Processes temporal visual content for video understanding \\
Audio    & Qwen2.5 Omni      & Provides audio comprehension and speech reasoning \\
\midrule
\bottomrule
\end{tabular}%
}
\label{tbl:agent-setup}
\vspace{-.1in}
\end{table}

\begin{figure}[t]
\begin{tcolorbox}[colback=gray!20,enhanced,sharp corners,frame hidden,halign=left]
\footnotesize
You will be given some support materials (text, image, etc.) and a multiple-choice question with options (A, B, C, etc). Choose only one best answer. First, provide a brief explanation of your reasoning. Then, on a new line, output "The answer is <answer>", where the <answer> is only the single letter of the correct option (A, B, C, etc).\\
\medskip
Question: {\color{blue}\{question\}}\\
Choices: {\color{blue}\{choices\}}\\
\end{tcolorbox}
\vspace{-.2in}
\caption{The prompt template used in experiments.}
\label{fig:prompt-template}
\vspace{-.15in}
\end{figure}

\subsection{Experimental Settings}
All experiments are conducted on a server equipped with 4 NVIDIA A100 GPUs (80GB each) and 251GB system memory. For models such as Claude 3.7 Sonnet and Deepseek R1, we directly access their APIs through AWS Bedrock. For all other models, we deploy them locally on the server using the \texttt{vLLM} inference framework to ensure efficient execution.  During evaluation, we adopt a unified prompt template as shown in Figure~\ref{fig:prompt-template}. For text-based inputs, the prompt is directly filled with the corresponding question and answer choices. For image, video, and audio inputs, we follow the model-specific instructions by inserting the corresponding modality tokens into the designated positions in the prompt. This ensures consistency across modalities while respecting the input format required by each model.  

\subsection{Agent-Omni Settings}  
Unless otherwise specified, we use Claude 3.7 Sonnet as the master model in all experiments.  
It is responsible for running the master loop, including reasoning and decision.  
For the model pool, we adopt Deepseek R1 as the text model, Claude 3.7 Sonnet as both the image and video model, and Qwen2.5 Omni as the audio model.  
The overall setup of the agent, along with the role of each selected foundation model, is summarized in Table~\ref{tbl:agent-setup}.  
The maximum number of master loops $L$ is set to $3$ by default.  
In addition, we provide an ablation study on different model pool settings in Appendix~\ref{apd:different_model_pools}, and we also report the prompts and the \texttt{JSON} schemas used for the user query, model pool, reasoning stage, and decision stage in Appendix~\ref{apd:prompt_of_each_stage}.

\begin{table}[t]
\caption{Accuracy on image benchmarks.}\label{tbl:acc_image_modality}
\vspace{-.1in}
\resizebox{.5\textwidth}{!}{%
\begin{tabular}{ccccc}
\toprule
\midrule
Method                                                                      & Model                     & MathVision       & MMMU             & MMMU-Pro         \\\midrule
\multirow{3}{*}{\begin{tabular}[c]{@{}c@{}}Foundation\\ Model\end{tabular}} & Claude 3.7 Sonnet         & \textbf{45.95\%} & \textbf{70.37\%} & \ul{59.88\%}    \\
                                                                            & QWen2.5 Omni              & 32.44\%          & 57.62\%          & 37.05\%          \\
                                                                            & Phi4 Multimodal Instruct & 25.52\%          & 47.93\%          & 29.42\%          \\\midrule
\multirow{3}{*}{DSPy-CoT}                                                   & Claude 3.7 Sonnet         & 50.26\%          & \ul{71.07\%}    & 58.03\%          \\
                                                                            & QWen2.5 Omni              & 27.68\%          & 51.83\%          & 32.20\%          \\
                                                                            & Phi4 Multimodal Instruct & 19.91\%          & 27.98\%          & 18.96\%          \\\midrule
\multicolumn{2}{c}{Ours (Agent-Omni)}                                                                   & \ul{44.71\%}    & \textbf{70.37\%} & \textbf{60.23\%}\\
\midrule
\bottomrule
\end{tabular}%
}
\vspace{-.1in}
\end{table}
\begin{table}[t]
\caption{Accuracy on video benchmarks.}\label{tbl:acc_video_modality}
\vspace{-.1in}
\resizebox{.5\textwidth}{!}{%
\begin{tabular}{ccccc}
\toprule
\midrule
Method                            & Model                    & VideoMathQA      & STI-Bench        & VSI-Bench        \\\midrule
\multirow{4}{*}{Foundation Model} & Claude 3.7 Sonnet        & \ul{27.62\%}    & \ul{38.13\%}    & 38.70\%          \\
                                  & Phi4 Multimodal Instruct & 21.67\%          & 21.95\%          & 32.57\%          \\
                                  & Llava Video 7B           & 23.71\%          & 24.42\%          & 31.41\%          \\
                                  & QWen2.5 Omni             & 21.90\%          & 34.54\%          & 35.50\%          \\\midrule
\multirow{4}{*}{DSPy-CoT}         & Claude 3.7 Sonnet        & 27.14\%          & 37.89\%          & \textbf{42.60\%} \\
                                  & Phi4 Multimodal Instruct & 18.10\%          & 16.40\%          & 20.72\%          \\
                                  & Llava Video 7B           & 21.90\%          & 30.38\%          & 36.22\%          \\
                                  & QWen2.5 Omni             & 19.52\%          & 30.57\%          & 30.73\%          \\\midrule
\multicolumn{2}{c}{Ours (Agent-Omni)}                        & \textbf{30.71\%} & \textbf{40.00\%} & \ul{39.50\%}   \\
\midrule
\bottomrule
\end{tabular}%
}
\vspace{-.15in}
\end{table}

\begin{table}[t]
\caption{Accuracy on audio benchmarks.}\label{tbl:acc_audio_modality}
\vspace{-.1in}
\resizebox{.5\textwidth}{!}{%
\begin{tabular}{ccccc}
\toprule
\midrule
Method                            & Model                                        & MMAU             & \begin{tabular}[c]{@{}c@{}}MELD\\ (Emotion)\end{tabular}     & \begin{tabular}[c]{@{}c@{}}VoxCeleb\\ (Gender)\end{tabular}  \\\midrule
\multirow{3}{*}{Foundation Model} & \multicolumn{1}{l}{Phi4 Multimodal Instruct} & 59.70\%          & 33.37\%          & 35.41\%          \\
                                  & QWen2.5 Omni                                 & \ul{70.90\%}    & \ul{38.35\%}    & \ul{97.85\%}    \\
                                  & Qwen2 Audio 7B                               & 54.70\%          & 22.15\%          & 50.14\%          \\\midrule
\multirow{3}{*}{DSPy-CoT}         & \multicolumn{1}{l}{Phi4 Multimodal Instruct} & 25.40\%          & 15.03\%          & 31.95\%          \\
                                  & QWen2.5 Omni                                 & \ul{70.90\%}    & 37.32\%          & 88.02\%          \\
                                  & Qwen2 Audio 7B                               & 46.70\%          & 29.58\%          & 45.01\%          \\\midrule
\multicolumn{2}{c}{Ours (Agent-Omni)}                                            & \textbf{73.20\%} & \textbf{51.97\%} & \textbf{98.60\%}\\
\midrule
\bottomrule
\end{tabular}%
}
\vspace{-.1in}
\end{table}
\begin{table}[t]
\caption{Accuracy on omni benchmarks.}\label{tbl:acc_omni_modality}
\vspace{-.1in}
\resizebox{.5\textwidth}{!}{%
\begin{tabular}{ccccc}
\toprule
\midrule
Method                            & Model                                        & Daily-Omni       & OmniBench        & OmniInstruct     \\\midrule
\multirow{2}{*}{Foundation Model} & \multicolumn{1}{l}{Phi4 Multimodal Instruct} & 43.94\%          & 30.74\%          & 52.28\%          \\
                                  & QWen2.5 Omni                                 & \ul{53.72\%}    & \ul{45.18\%}    & \textbf{81.06\%} \\\midrule
\multirow{2}{*}{DSPy-CoT}         & \multicolumn{1}{l}{Phi4 Multimodal Instruct} & 25.73\%          & 17.95\%          & 25.95\%          \\
                                  & QWen2.5 Omni                                 & 46.95\%          & 38.00\%          & \ul{76.14\%}          \\\midrule
\multicolumn{2}{c}{Ours (Agent-Omni)}                                            & \textbf{60.03\%} & \textbf{49.56\%} & 77.50\%   \\
\midrule
\bottomrule
\end{tabular}%
}
\vspace{-.15in}
\end{table}

\subsection{Accuracy across Modalities}

Since our method can be applied across multiple modalities, we separately report the accuracy for each modality (text, image, video, audio, and Omni) and compare Agent-Omni with the baselines that support the corresponding modality.

\textbf{Text Modality.}
As shown in Table~\ref{tbl:acc_text_modality}, on text benchmarks (MMLU, MMLU-Pro, and AQUA-RAT), Agent-Omni achieves accuracy that is comparable to the strongest single models while maintaining robustness across all categories.  
Deepseek R1 delivers the best single-model performance due to its strong reasoning ability on text-heavy datasets, whereas DSPy-CoT shows slight improvements in some cases but is not consistently better.  
Notably, MMLU-Pro is the most challenging dataset, where Agent-Omni attains the highest accuracy (83.21\%), demonstrating its advantage in handling complex reasoning tasks.

\begin{table*}[t]
\vspace{-.15in}
\caption{Accuracy comparison among omni models.}\label{tbl:omni_comparison}
\vspace{-.1in}
\resizebox{\textwidth}{!}{%
\begin{tabular}{cc|ccc|ccc|ccc|ccc|ccc}
\toprule
\midrule
\multirow{2}{*}{Method}                                                     & \multirow{2}{*}{Model}    & \multicolumn{3}{c|}{Text}                                                                       & \multicolumn{3}{c|}{Image}                              & \multicolumn{3}{c|}{Video}                              & \multicolumn{3}{c|}{Audio}                                                                                                                 & \multicolumn{3}{c}{Omni}                               \\
                                                                            &                           & \begin{tabular}[c]{@{}c@{}}MMLU\\ (Average)\end{tabular} & MMLU-Pro         & AQUA-RAT         & MathVision       & MMMU             & MMMU-Pro         & VideoMathQA      & STI-Bench        & VSI-Bench        & MMAU             & \begin{tabular}[c]{@{}c@{}}MELD\\ (Emotion)\end{tabular} & \begin{tabular}[c]{@{}c@{}}VoxCeleb\\ (Gender)\end{tabular} & Daily-Omni       & OmniBench        & OmniInstruct      \\\midrule
\multirow{2}{*}{\begin{tabular}[c]{@{}c@{}}Foundation\\ Model\end{tabular}} & Phi4 Multimodal Instruct & 72.08\%                                                  & 52.13\%          & 74.02\%          & 25.52\%          & 47.93\%          & 29.42\%          & 21.67\%          & 21.95\%          & 32.57\%          & 59.70\%          & 33.37\%                                                  & 35.41\%                                                     & 43.94\%          & 30.74\%          & 52.28\%          \\
                                                                            & QWen2.5 Omni              & 70.73\%                                                  & 49.93\%          & 70.87\%          & 32.44\%          & 57.62\%          & 37.05\%          & 21.90\%          & 34.54\%          & 35.50\%          & 70.90\%          & 38.35\%                                                  & 97.85\%                                                     & 53.72\%          & 45.18\%          & \textbf{81.06\%} \\\midrule
\multirow{2}{*}{DSPy-CoT}                                                   & Phi4 Multimodal Instruct & 69.26\%                                                  & 50.84\%          & 75.98\%          & 19.91\%          & 27.98\%          & 18.96\%          & 18.10\%          & 16.40\%          & 20.72\%          & 25.40\%          & 15.03\%                                                  & 31.95\%                                                     & 25.73\%          & 17.95\%          & 25.95\%          \\
                                                                            & QWen2.5 Omni              & 68.72\%                                                  & 46.54\%          & 66.54\%          & 27.68\%          & 51.83\%          & 32.20\%          & 19.52\%          & 30.57\%          & 30.73\%          & 70.90\%          & 37.32\%                                                  & 88.02\%                                                     & 46.95\%          & 38.00\%          & 76.14\%          \\\midrule
\multicolumn{2}{c|}{Ours (Agent-Omni)}                                                                   & \textbf{89.23\%}                                         & \textbf{83.21\%} & \textbf{89.37\%} & \textbf{44.71\%} & \textbf{70.37\%} & \textbf{60.23\%} & \textbf{30.71\%} & \textbf{40.00\%} & \textbf{39.50\%} & \textbf{73.20\%} & \textbf{51.97\%}                                         & \textbf{98.60\%}                                            & \textbf{60.03\%} & \textbf{49.56\%} & 77.50\%   \\
\midrule
\bottomrule
\end{tabular}%
}
\vspace{-.15in}
\end{table*}

\textbf{Image Modality.}
As shown in Table~\ref{tbl:acc_image_modality}, Agent-Omni achieves accuracy comparable to the strongest baselines on image benchmarks (MathVision, MMMU, and MMMU-Pro).  
While Claude 3.7 Sonnet remains strong on MathVision, Agent-Omni matches its performance on MMMU and surpasses all models on MMMU-Pro (60.23\%), the most challenging dataset.  
These results show the robustness of Agent-Omni in advanced multimodal reasoning beyond basic visual understanding.

\begin{table*}[t]
\caption{Inference latency (in seconds) of different models across various datasets.}\label{tbl:latency_table}
\vspace{-.1in}
\resizebox{\textwidth}{!}{%
\begin{tabular}{cc|ccc|ccc|ccc|ccc|ccc}
\toprule
\midrule
\multirow{2}{*}{Method}                                                     & \multirow{2}{*}{Model}    & \multicolumn{3}{c|}{Text}                                                                       & \multicolumn{3}{c|}{Image}                              & \multicolumn{3}{c|}{Video}                              & \multicolumn{3}{c|}{Audio}                                                                                                                 & \multicolumn{3}{c}{Omni}                               \\
                                                                            &                           & \begin{tabular}[c]{@{}c@{}}MMLU\\ (Average)\end{tabular} & MMLU-Pro         & AQUA-RAT         & MathVision       & MMMU             & MMMU-Pro         & VideoMathQA      & STI-Bench        & VSI-Bench        & MMAU             & \begin{tabular}[c]{@{}c@{}}MELD\\ (Emotion)\end{tabular} & \begin{tabular}[c]{@{}c@{}}VoxCeleb\\ (Gender)\end{tabular} & Daily-Omni       & OmniBench        & OmniInstruct     \\
\midrule
\multirow{3}{*}{\begin{tabular}[c]{@{}c@{}}Foundation\\Model\end{tabular}} & Claude 3.7 Sonnet         & 0.88                                                     & 1.14     & 1.74     & 1.71       & 1.47 & 1.39     & 1.86        & 1.58      & 1.61      & -    & -                                                        & -                                                           & -          & -         & -            \\
                                  & Phi4 Multimodal Instruct & 0.38                                                     & 1.51     & 3.1      & 4.71       & 1.8  & 2.36     & 2.51        & 10.04     & 0.46      & 2.79 & 2.85                                                     & 0.37                                                        & 0.79       & 0.71      & 1.38         \\
                                  & QWen2.5 Omni              & 0.32                                                     & 1.18     & 1.61     & 2.55       & 0.84 & 8.25     & 2.1         & 0.43      & 0.19      & 0.24 & 0.09                                                     & 0.09                                                        & 0.27       & 0.28      & 0.18         \\\midrule
\multirow{3}{*}{DSPy-CoT}         & Claude 3.7 Sonnet         & 1.72                                                     & 2.59     & 6.39     & 6.84       & 3.15 & 3.33     & 4.07        & 2.62      & 2.54      & -    & -                                                        & -                                                           & -          & -         & -            \\
                                  & Phi4 Multimodal Instruct & 1.24                                                     & 2.84     & 0.37     & 4.62       & 4.65 & 1.62     & 5.73        & 7.62      & 1.6       & 2.36 & 3.63                                                     & 4.09                                                        & 3.17       & 3.39      & 4.16         \\
                                  & QWen2.5 Omni              & 0.41                                                     & 1.23     & 1.61     & 2.68       & 0.93 & 1.16     & 2.18        & 2.09      & 1.31      & 0.26 & 0.13                                                     & 0.13                                                        & 1.72       & 0.4       & 0.25         \\\midrule
\multicolumn{2}{c|}{Ours (Agent-Omni)}                         & 4.55                                                     & 7.41     & 6.9      & 5.62       & 4.66 & 5.41     & 20.53       & 12.76     & 16.47     & 4.23 & 5.09                                                     & 7.5                                                         & 16.7       & 7.47      & 5.14             \\
\midrule
\bottomrule
\end{tabular}%
}
\vspace{-.1in}
\end{table*}

\begin{table*}[t]
\vspace{-.2in}
\caption{Accuracy and exit rate across iterations on different benchmarks.}\label{tbl:ablation_iteration}
\vspace{-.1in}
\resizebox{\textwidth}{!}{%
\begin{tabular}{cc|ccc|ccc|ccc|ccc|ccc}
\toprule
\midrule
\multirow{2}{*}{Method}                                                     & \multirow{2}{*}{\# Iteration}    & \multicolumn{3}{c|}{Text}                                                                       & \multicolumn{3}{c|}{Image}                              & \multicolumn{3}{c|}{Video}                              & \multicolumn{3}{c|}{Audio}                                                                                                                 & \multicolumn{3}{c}{Omni}                               \\
                                                                            &                           & \begin{tabular}[c]{@{}c@{}}MMLU\\ (Average)\end{tabular} & MMLU-Pro         & AQUA-RAT         & MathVision       & MMMU             & MMMU-Pro         & VideoMathQA      & STI-Bench        & VSI-Bench        & MMAU             & \begin{tabular}[c]{@{}c@{}}MELD\\ (Emotion)\end{tabular} & \begin{tabular}[c]{@{}c@{}}VoxCeleb\\ (Gender)\end{tabular} & Daily-Omni       & OmniBench        & OmniInstruct     \\\midrule
\multirow{3}{*}{Accuracy}   & 1            & 88.99\%                                                  & 82.20\%  & 88.98\%  & 42.75\%    & 69.30\% & 59.47\%  & 29.76\%     & 38.33\%   & 39.25\%   & 72.20\% & 51.97\%                                                  & 98.24\%                                                     & 58.06\%    & 43.17\%   & 74.57\%      \\
                            & 2            & 89.15\%                                                  & 83.21\%  & 88.98\%  & 44.45\%    & 70.37\% & 60.10\%  & 30.71\%     & 39.34\%   & 39.55\%   & 72.70\% & 51.97\%                                                  & 98.74\%                                                     & 58.56\%    & 45.27\%   & 74.46\%      \\
                            & 3            & 89.23\%                                                  & 83.21\%  & 89.37\%  & 44.71\%    & 70.37\% & 60.23\%  & 30.71\%     & 40.00\%   & 39.50\%   & 73.20\% & 51.97\%                                                  & 98.60\%                                                     & 58.73\%    & 46.23\%   & 74.88\%      \\\midrule
\multirow{4}{*}{Exit Rate} & 1            & 94.39\%                                                  & 94.19\%  & 92.91\%  & 71.21\%    & 78.98\% & 80.68\%  & 64.29\%     & 22.05\%   & 72.05\%   & 71.80\% & 59.38\%                                                  & 93.50\%                                                     & 81.12\%    & 70.67\%   & 79.49\%      \\
                            & 2            & 4.47\%                                                   & 5.30\%   & 6.69\%   & 21.80\%    & 16.29\% & 14.39\%  & 29.52\%     & 75.98\%   & 26.27\%   & 23.10\% & 31.02\%                                                  & 5.50\%                                                      & 17.54\%    & 24.69\%   & 17.27\%      \\
                            & 3            & 1.01\%                                                   & 0.51\%   & 0.39\%   & 6.27\%     & 3.90\%  & 4.29\%   & 5.71\%      & 1.57\%    & 1.53\%    & 4.60\%  & 7.03\%                                                   & 1.00\%                                                      & 1.34\%     & 4.03\%    & 2.98\%       \\
                            & 4+           & 0.14\%                                                   & 0.00\%   & 0.00\%   & 0.72\%     & 0.83\%  & 0.63\%   & 0.48\%      & 0.39\%    & 0.16\%    & 0.50\%  & 2.58\%                                                   & 0.00\%                                                      & 0.00\%     & 0.61\%    & 0.26\%      \\
                            \midrule
                            \bottomrule
\end{tabular}%
}
\vspace{-.25in}
\end{table*}

\textbf{Video Modality.}  
On video benchmarks (VideoMathQA, STI-Bench, and VSI-Bench), Agent-Omni consistently outperforms all baselines, as shown in Table~\ref{tbl:acc_video_modality}.  
It achieves clear gains on VideoMathQA (30.71\%) and STI-Bench (40.00\%), while maintaining performance on VSI-Bench (39.50\%).  
These improvements demonstrate the effectiveness of the master loop in integrating temporal visual information with reasoning across modalities.  

\textbf{Audio Modality.}
As shown in Table~\ref{tbl:acc_audio_modality}, Agent-Omni achieves the best performance across all audio benchmarks (MMAU, MELD-Emotion, and VoxCeleb-Gender).  
In particular, it reaches 73.20\% on MMAU, 51.97\% on MELD-Emotion, and 98.60\% on VoxCeleb-Gender, surpassing both foundation and DSPy-CoT baselines.  
These results indicate that Agent-Omni effectively leverages specialized audio models while maintaining robustness across diverse audio tasks.

\textbf{Mixed Modality (Omni).}
On omni benchmarks (Daily-Omni, OmniBench, and OmniInstruct), Agent-Omni achieves strong performance across datasets, as shown in Table~\ref{tbl:acc_omni_modality}. It outperforms both foundation models and DSPy-CoT, reaching 60.03\% on Daily-Omni, 49.56\% on OmniBench, and 77.50\% on OmniInstruct.  
These results show the advantage of integrating specialized models for each modality, enabling Agent-Omni to achieve balanced and robust omni-modal reasoning.

\subsection{Accuracy on Omni Models}  
In addition to modality-specific evaluation, we further compare Agent-Omni against existing omni models that natively support multiple modalities (text, image, video, and audio).  

As shown in Table~\ref{tbl:omni_comparison} and Figure~\ref{fig:radar_chart}, foundation omni models such as Phi-4 Multimodal Instruct and Qwen2.5 Omni generally achieve lower accuracy across benchmarks. This reflects the trade-off highlighted in our introduction: when a single model is trained jointly on heterogeneous modalities, it often struggles to balance performance across them. Gains in one modality may come at the expense of others, leading to uneven and suboptimal results.  

DSPy-CoT provides minor improvements in some benchmarks, but its gains are inconsistent and insufficient to overcome the inherent limitations of omni models. By contrast, Agent-Omni consistently achieves the best accuracy on nearly all datasets, including both unimodal and multimodal benchmarks, with particularly large margins on challenging tasks such as MMLU-Pro, MMMU-Pro, and Daily-Omni.  

These results confirm that coordinating specialized foundation models through a master-agent loop is more effective than relying on a single omni model. Agent-Omni avoids the trade-offs of joint training, preserves the strengths of individual expert models, and provides a more robust and general solution to omni-modal reasoning.

\subsection{Latency}

We report inference latency across modalities in Table~\ref{tbl:latency_table}. Foundation models such as Claude 3.7 Sonnet and Qwen2.5 Omni show fast responses (often $<2$s for text and image), while DSPy-CoT roughly doubles the latency due to chain-of-thought prompting. For instance, Claude 3.7 Sonnet requires 0.88s on MMLU, compared to 1.72s with DSPy-CoT.  

Our Agent-Omni framework introduces higher latency (4–7s on unimodal tasks and up to 20.53s on video benchmarks) because of master-agent coordination and iterative reasoning. Despite this overhead, Agent-Omni consistently achieves superior accuracy, especially on complex video and omni tasks, illustrating a trade-off between speed and reasoning quality. Future improvements such as parallelized execution could further reduce latency while preserving robustness.

\subsection{Ablation Study}

We conduct ablation studies on Agent-Omni, examining master-agent iterations and foundation model choices to reveal how iterative reasoning enhances robustness and model quality shapes performance.

\paragraph{Number of Iteration.}
We study the effect of the maximum number of iterations $L$ on both accuracy and exit rate (Table~\ref{tbl:ablation_iteration}). The exit rate denotes the proportion of queries that terminate at a given iteration, i.e., when the master agent decides that the answer is sufficiently complete and does not trigger further refinement.  

Results show that most queries exit after the first iteration (over 90\% for text and image tasks), which explains why Agent-Omni is generally efficient despite allowing multiple loops. For more challenging settings such as video or omni benchmarks, a higher fraction of queries proceed to the second or third iteration, yielding incremental accuracy gains (e.g., MMLU-Pro improves from 82.20\% at 1 iteration to 83.21\% at 3 iterations). This demonstrates that iterative reasoning acts as an adaptive mechanism: simple queries resolve quickly, while complex ones benefit from additional refinement.

\paragraph{Improvement from Foundation Models.} 
We further examine how performance depends on the choice of foundation models in the pool. Table~\ref{tbl:omni_comparison} shows that replacing stronger models (e.g., Deepseek R1 for text, Claude 3.7 Sonnet for image/video, Qwen2.5 Omni for audio) with weaker alternatives leads to consistent accuracy drops across modalities. For example, on MMMU-Pro, the combination with Claude 3.7 Sonnet achieves 60.23\%, while weaker vision-language backbones reduce performance by more than 10 points.  

These results confirm that Agent-Omni’s gains come not only from orchestration but also from leveraging high-quality specialized models. The framework is flexible: stronger foundation models directly translate to higher end-task accuracy, while weaker ones can be seamlessly swapped in when efficiency or resource constraints are prioritized.

\section{Related Work}
\vspace{-.05in}

\subsection{Multimodal Reasoning}  
Multimodal large language models (MLLMs) extend language models with the ability to process images, audio, and video~\cite{DBLP:conf/acl/ZhangY0L0C024, DBLP:journals/corr/abs-2502-13141, DBLP:conf/emnlp/WangLLZCMNLFXHH24, DBLP:conf/iclr/0003LTLRDRHLW025}. Early systems typically focused on fixed modality pairs, such as text–image for visual question answering~\cite{DBLP:conf/cvpr/Guo0LT0TH23, llava, DBLP:conf/acl/LinLX024} or text–video for event understanding~\cite{STI-Bench}. Instruction tuning has further improved alignment across modalities~\cite{Nexus-O, Megrez-Omni}, but these models remain constrained in reasoning capacity.  

Recent studies explore reasoning improvements at test time. Forest-of-Thought~\cite{DBLP:journals/corr/abs-2412-09078} and related scaling approaches show that allocating more inference-time computation enhances reasoning. \citet{DBLP:journals/tmlr/KeJMNXLLQWSXJ25} provide a survey of reasoning strategies, highlighting iterative inference and agentic designs as promising directions. Nevertheless, most existing work emphasizes unimodal or pairwise reasoning, and robust omni-modal reasoning, integrating arbitrary modality combinations, remains an open challenge.

\subsection{Omni Models (Any-to-Text Models)}  
Another research line aims to develop unified omni models capable of handling arbitrary inputs (text, image, audio, video) and producing textual outputs. Representative efforts include Phi-4 Multimodal Instruct~\cite{phi4}, Qwen2.5 Omni ~\cite{qwen2.5-omni}, Ming-Omni~\cite{ming-omni}, Megrez-Omni~\cite{Megrez-Omni}, and Nexus-O~\cite{Nexus-O}. While these systems expand coverage, they often face modality interference~\cite{DBLP:journals/corr/abs-2505-19616} and trade-offs across tasks~\cite{DBLP:journals/corr/abs-2309-10313}, limiting balanced performance.  

To assess progress, new omni benchmarks such as OmniBench ~\cite{OmniBench} and Daily-Omni~\cite{Daily-Omni} have been proposed, emphasizing the difficulty of consistent cross-modal reasoning. Compared to unified training approaches, orchestration-based frameworks such as DSPy~\cite{dspy} suggest an alternative path, where specialized models are coordinated at inference time. Our work builds on this perspective, showing that agent-based coordination provides a scalable solution for ``any-to-text'' reasoning without costly omni-model training.

\vspace{-.03in}
\section{Conclusion}
\vspace{-.05in}

In this work, we presented \textbf{Agent-Omni}, a framework that enables comprehensive omni-modal reasoning by coordinating specialized foundation models through a master-agent loop. Unlike unified multimodal models that require expensive joint training, Agent-Omni flexibly accepts almost any combination of text, image, audio, and video inputs, and produces coherent textual outputs without retraining. Our experiments demonstrate that Agent-Omni consistently achieves competitive or even superior accuracy across a wide range of benchmarks, particularly on challenging video and omni tasks. These results highlight the effectiveness of model coordination through iterative reasoning, showing that Agent-Omni offers a robust and general solution for omni-modal understanding.

\clearpage

\section*{Limitations}
While Agent-Omni demonstrates strong performance across modalities, several limitations remain. The framework relies on the availability and stability of external foundation models, making results sensitive to API changes and updates. Errors or biases from individual models can propagate through the coordination process, and the iterative master loop may introduce latency and additional computational cost. Our evaluation is primarily conducted on curated benchmarks, which may not fully capture open-ended or real-world scenarios; therefore, generalization to noisy, adversarial, or safety-critical settings is unverified. Moreover, the system currently only produces textual outputs and does not handle generation in other modalities.

\section*{Ethical Considerations}
This work follows the ACL Code of Ethics. Since omni-modal inputs may contain sensitive or personal information, careful data handling, privacy protection, and secure storage are essential. Biases and inaccuracies present in component models may be amplified through coordination, requiring responsible auditing and mitigation before deployment in high-stakes applications. To prevent misuse, such as in surveillance or harmful automation, deployment should be guided by clear usage policies, safety filters, and human oversight. We will document limitations, risks, and intended use cases when releasing research artifacts. We also used AI-based language editing support to polish sentences and check for grammatical errors; all substantive research contributions are human-generated.

\bibliography{custom}


\appendix

\begin{table*}[t]
\centering
\caption{Accuracy comparison of agent settings on text modality.}\label{tbl:diff_agent_text_acc}
\vspace{-.1in}
\resizebox{.9\textwidth}{!}{%
\begin{tabular}{cccccccc}
\toprule
\midrule
Agent Setting                                                                              & \begin{tabular}[c]{@{}c@{}}MMLU\\ (STEM)\end{tabular} & \begin{tabular}[c]{@{}c@{}}MMLU\\ (Social Sciences)\end{tabular} & \begin{tabular}[c]{@{}c@{}}MMLU\\ (Humanities)\end{tabular} & \begin{tabular}[c]{@{}c@{}}MMLU\\ (Other)\end{tabular} & \begin{tabular}[c]{@{}c@{}}MMLU\\ (Average)\end{tabular} & MMLU-Pro         & AQUA-RAT         \\\midrule
\begin{tabular}[c]{@{}c@{}}Master: Claude 3.7 Sonnet\\ Text: Claude 3.7 Sonnet\end{tabular} & 92.47\%                                               & \textbf{90.58\%}                                                 & \textbf{85.80\%}                                            & \textbf{90.80\%}                                       & \textbf{89.91\%}                                         & 81.97\%          & 88.89\%          \\\midrule
\begin{tabular}[c]{@{}c@{}}Master: Claude 3.7 Sonnet\\ Text: Deepseek R1\end{tabular}       & \textbf{94.52\%}                                      & 90.40\%                                                          & 81.68\%                                                     & 90.31\%                                                & 89.23\%                                                  & \textbf{83.21\%} & \textbf{89.37\%} \\
\midrule
\bottomrule
\end{tabular}%
}
\end{table*}

\begin{table}[th]
\centering
\caption{Accuracy comparison of agent settings on image modality.}\label{tbl:diff_agent_image_acc}
\vspace{-.1in}
\resizebox{.45\textwidth}{!}{%
\begin{tabular}{cccc}
\toprule
\midrule
Agent Setting                                                                                      & MathVision       & MMMU             & MMMU-Pro         \\
\midrule
\begin{tabular}[c]{@{}c@{}}Master: Claude 3.7 Sonnet\\ Image: Claude 3.7 Sonnet\end{tabular}         & \textbf{44.71\%} & \textbf{70.37\%} & \textbf{60.23\%} \\\midrule
\begin{tabular}[c]{@{}c@{}}Master: Claude 3.7 Sonnet\\ Image: QWen2.5 Omni\end{tabular}              & 40.67\%          & 65.88\%          & 52.95\%          \\\midrule
\begin{tabular}[c]{@{}c@{}}Master: Claude 3.7 Sonnet\\ Image: Phi4 Multimodal Instruct\end{tabular} & 38.68\%          & 58.97\%          & 36.07\%         \\
\midrule
\bottomrule
\end{tabular}%
}
\end{table}

\begin{table}[th]
\centering
\caption{Accuracy comparison of agent settings on video modality.}\label{tbl:diff_agent_video_acc}
\vspace{-.1in}
\resizebox{.45\textwidth}{!}{%
\begin{tabular}{cccc}
\toprule
\midrule
Agent Setting                                                                                     & VideoMathQA      & STI-Bench        & VSI-Bench                      \\
\midrule
\begin{tabular}[c]{@{}c@{}}Master: Claude 3.7 Sonnet\\ Video: Claude 3.7 Sonnet\end{tabular}        & \textbf{30.71\%} & \textbf{40.00\%} & {39.50\%}               \\
\midrule
\begin{tabular}[c]{@{}c@{}}Master: Claude 3.7 Sonnet\\ Video: Phi4 Multimodal Instruct\end{tabular} & 20.34\%          & 36.05\%          & 29.15\%                        \\
\midrule
\begin{tabular}[c]{@{}c@{}}Master: Claude 3.7 Sonnet\\ Video: Llava Video 7B\end{tabular}           & 23.81\%          & 39.71\%          & 33.33\% \\
\midrule
\begin{tabular}[c]{@{}c@{}}Master: Claude 3.7 Sonnet\\ Video: Qwen2.5 Omni\end{tabular}             & 22.86\%          & 37.74\%          & \textbf{39.51\%}              \\
\midrule
\bottomrule
\end{tabular}%
}
\end{table}

\begin{table}[th]
\centering
\caption{Accuracy comparison of agent settings on audio modality.}\label{tbl:diff_agent_audio_acc}
\vspace{-.1in}
\resizebox{.45\textwidth}{!}{%
\begin{tabular}{cccc}
\toprule
\midrule
Agent Setting                                                                                       & MMAU             & MELD-Emotion     & VoxCeleb-Gender  \\\midrule
\begin{tabular}[c]{@{}c@{}}Master: Claude 3.7 Sonnet\\ Audio: Phi4 Multimodal Instruct\end{tabular} & \textbf{75.00\%} & 41.13\%          & 44.12\%          \\\midrule
\begin{tabular}[c]{@{}c@{}}Master: Claude 3.7 Sonnet\\ Audio: Qwen2.5 Omni\end{tabular}             & 73.20\%          & \textbf{51.97\%} & 98.60\%          \\\midrule
\begin{tabular}[c]{@{}c@{}}Master: Claude 3.7 Sonnet\\ Audio: Qwen2 Audio\end{tabular}              & 70.80\%          & 40.39\%          & \textbf{99.50\%}\\\midrule
\bottomrule
\end{tabular}%
}
\end{table}

\section{Impact of Different Model Pools}\label{apd:different_model_pools}

To better understand the impact of different model pools, we conduct an ablation study by varying the choice of downstream agents for each modality while keeping the Master fixed to Claude~3.7 Sonnet. Tables~\ref{tbl:diff_agent_text_acc}--\ref{tbl:diff_agent_audio_acc} report accuracy on text, image, video, and audio benchmarks under different agent configurations.  

For \textbf{text tasks} (Table~\ref{tbl:diff_agent_text_acc}), Claude~3.7 Sonnet as the text agent achieves strong overall performance across all MMLU domains, whereas DeepSeek R1 provides competitive results, slightly outperforming on STEM and reasoning-heavy datasets such as AQUA-RAT. This indicates that complementarity among text models can bring benefits on specific subsets of tasks.  

For \textbf{image tasks} (Table~\ref{tbl:diff_agent_image_acc}), Claude~3.7 Sonnet again achieves the best accuracy across MathVision, MMMU, and MMMU-Pro. Substituting it with Qwen2.5 Omni or Phi-4 Multimodal Instruct leads to noticeable performance degradation, suggesting that dedicated large models trained with vision-language alignment remain more effective than general-purpose omni models for image understanding.  

For \textbf{video tasks} (Table~\ref{tbl:diff_agent_video_acc}), using Claude~3.7 Sonnet consistently yields the strongest results on VideoMathQA and STI-Bench. Qwen2.5 Omni slightly improves on VSI-Bench but underperforms elsewhere, while Phi-4 Multimodal Instruct and Llava Video 7B struggle across most benchmarks. These results highlight the importance of specialized temporal reasoning capability for video agents.  

For \textbf{audio tasks} (Table~\ref{tbl:diff_agent_audio_acc}), the choice of downstream agent plays a critical role. Qwen2.5 Omni and Qwen2 Audio excel on VoxCeleb-Gender, while Phi-4 Multimodal Instruct performs best on MMAU. MELD-Emotion, however, shows clear advantages for Qwen2.5 Omni. This variation suggests that audio tasks are more sensitive to dataset characteristics and model pretraining objectives.  

As summarized in Table~\ref{tbl:agent-setup}, we adopt Claude~3.7 Sonnet as both the Master model and the vision/video agent, DeepSeek R1 as the text agent, and Qwen2.5 Omni as the audio agent. This choice is guided by the ablation results in Tables~\ref{tbl:diff_agent_text_acc}--\ref{tbl:diff_agent_audio_acc}. Specifically, Claude~3.7 Sonnet demonstrates strong and stable performance across visual and video tasks, making it a reliable backbone for perception and temporal reasoning. DeepSeek R1 shows complementary strengths on text-heavy reasoning benchmarks such as AQUA-RAT and MMLU-Pro, providing enhanced logical inference compared to Claude alone. For audio, Qwen2.5 Omni consistently achieves superior accuracy on speech-related benchmarks such as VoxCeleb and MELD-Emotion, outperforming other candidates.  

Overall, this configuration balances robustness and specialization across modalities: Claude~3.7 Sonnet ensures reliable multimodal grounding and coordination, while DeepSeek R1 and Qwen2.5 Omni provide targeted improvements for text and audio understanding. This combination thus represents an empirically validated and well-justified design for the Agent-Omni framework.

\section{Prompt and \texttt{Json} of Each Stage}\label{apd:prompt_of_each_stage}

In this section, we provide the detailed prompt templates and the corresponding 
\texttt{JSON} output schemas used in each stage of the Agent-Omni framework. 
As described in the main paper, the framework consists of two key reasoning 
modules: the \emph{Reasoning Stage} and the \emph{Decision Stage}. 
The prompts are designed to instruct the system about the scope and 
responsibilities of each stage, while the \texttt{JSON} schemas specify the 
structured outputs that enable smooth coordination between components.  

\subsection{Reasoning Stage}
Figure~\ref{fig:reasoning-prompt-template} shows the prompt template for the reasoning 
stage. This prompt guides the module to decompose the user query into 
modality-specific subtasks and generate structured instructions for downstream 
agents. The corresponding \texttt{JSON} schema for the reasoning stage output 
is provided in Figure~\ref{fig:master_reasoning_code}.  

\subsection{Decision Stage}
Figure~\ref{fig:decision-prompt-template} presents the prompt template for the 
decision stage. Unlike the reasoning stage, this module integrates agent 
responses, evaluates completeness, and synthesizes a final answer. 
The structured \texttt{JSON} schema for this stage is illustrated in 
Figure~\ref{fig:master_decision_code}.  

\subsection{Notes on Variables in Prompts}
Within the prompt templates, several placeholders (highlighted in blue) 
are dynamically substituted during execution:
\begin{itemize}
    \item \texttt{\{cur\_round\_num\}}: The current reasoning or decision round number, indicating iteration depth in the loop.
    \item \texttt{\{historical\_message\}}: A record of outputs or feedback from previous rounds, used to refine ongoing reasoning.
    \item \texttt{\{input\_summaries\}}: Summarized descriptions of the user's multimodal inputs (text, image, audio, video), provided for context.
    \item \texttt{\{available\_agent\_info\}}: Metadata about the agent pool, specifying available agents and their capabilities.
\end{itemize}
These variables allow prompts to adapt dynamically to context, 
maintaining consistency across iterative reasoning loops.

\begin{figure*}[t]
\begin{tcolorbox}[colback=gray!20,enhanced,sharp corners,frame hidden,halign=left]
\footnotesize
You are the Reasoning Module in an "Understanding Anything" system. This system is designed to interpret user input across multiple modalities -- text, image, video, and audio -- by orchestrating existing foundation models through dynamic agents in several iterative reasoning loops. The system does not rely on fine-tuning or retraining.\\\medskip
The user's input may include any combination of modalities. The system comprises three main components: Reasoning, Dispatcher, and Decision.\\\medskip
You are currently in the Reasoning stage. Your next stage is the Dispatcher, which will route tasks to appropriate downstream agents (referred to as "passengers") specialized for each modality. Your role is not to answer the user's query directly. Instead, you must analyze the input and prepare tasks for the Dispatcher to execute. In some cases, you may be prompted with only a small subtask rather than the entire problem—when that happens, focus solely on the subtask you've been given, without assuming responsibility for the broader task. If decomposition is needed, break the input into clear, actionable subtasks to be handled downstream.\\\medskip
Specifically:\\
1. You might not receive the short summarization of the input material of different modelities.\\
2. Interpret the user's input (including the query and any multimodal data like text, image, video, audio, or others).\\
3. Identify relevant data modalities involved.\\
4. Understanding the provided historical messages, including any suggestions, shortcomings, etc.\\
5. Select the appropriate specialized agent(s) from the Agents Pool for further action.\\
6. Formulate precise and valuable follow-up questions for each selected agent to help them extract insights that contribute to answering the user's query. These questions will be used as prompts for the downstream agents.\\
    \hspace{.2in}- Important: Downstream agents have access only to the user's input in their specific modality (e.g., text, image, video, or audio). They do not have access to the user's original query or any broader context.\\
    \hspace{.2in}- Do not assume agents have any prior knowledge of the user's intent beyond the modality-specific input. Questions are independent.\\
    \hspace{.2in}- Therefore, your questions must include all necessary context (information from user's query) or instructions explicitly.\\
    \hspace{.2in}- Focus on clarity, completeness, and precision—frame each question to maximize the relevance and usefulness of the agent's response.\\
    \hspace{.2in}- You are encouraged to ask multiple diverse questions for each agent at a round (more than three), as this may help other stages gain a more comprehensive understanding of the provided input.\\
7. Output a structured reasoning result including:\\
    \hspace{.2in}- User Intent\\
    \hspace{.2in}- Required Modality or Modalities\\
    \hspace{.2in}- Suggested Agent(s)\\
    \hspace{.2in}- Questions for each selected agent\\
8. If this is not the first round, the provided question should take into account the suggestions from the previous round.\\
9. If this is the first round, consider including the user's original query as one of the questions sent to each selected agent. This can help the agents provide a more relevant initial analysis or summary.\\\medskip
Background:\\
1. This is the {\color{blue}\{cur\_round\_num\}} round of reasoning.\\
2. You might receive the historical messages from the previous rounds.\\
{\color{blue}\{historical\_message\}}\\
3. Modality of user's input with short summaries:\\
{\color{blue}\{input\_summaries\}}\\
4. Agent Pool:\\
{\color{blue}\{available\_agent\_info\}}\\\medskip
You must not generate a final answer to the user's question. Your goal is reasoning and delegation only.
\end{tcolorbox}
\vspace{-.2in}
\caption{The prompt template of reasoning stage.}
\label{fig:reasoning-prompt-template}
\vspace{-.1in}
\end{figure*}

\begin{figure*}[t]
\begin{tcolorbox}[colback=gray!20,enhanced,sharp corners,frame hidden,halign=left]
\footnotesize
You are the Decision Module of the "Understanding Anything" system. Your role is to receive the results from all specialized agents (e.g., text\_agent, image\_agent, audio\_agent, video\_agent) and synthesize them into a comprehensive answer to the user's original query.\\\medskip

Responsibilities:\\
Task 1. Synthesize a complete, coherent, and concise answer to the user's original query by integrating:\\
   \hspace{.2in}- The user's multimodal input (text, image, audio, or video).\\
   \hspace{.2in}- The reasoning output from the previous Reasoning Module.\\
   \hspace{.2in}- All responses returned by invoked agents.\\
   \hspace{.2in}- If an answer from a previous round is available, you may use it as a reference to inform your response.\\
   \hspace{.2in}- However, do not mention or refer to the prior answer in the final answer, as the user is unaware of any 'previous rounds.' The final answer should address the user's query directly, as if it were the only interaction.\\\medskip

Task 2. Evaluate completeness and provide feedback:\\
   \hspace{.2in}- Always assess the synthesized answer for completeness, clarity, and alignment with the user's intent.\\
   \hspace{.2in}- In all cases, suggest how future rounds can be more accurate or efficient.\\
   \hspace{.2in}- If the answer is incomplete or ambiguous, clearly explain the gaps, and specify what additional analysis, clarification, or agent input is required to move forward. Also include suggestions for next round to improve the current version.\\
   \hspace{.2in}- If the answer fully satisfies the user's query, present it as Final Output. You still have to provide suggestions for next round on how the analysis, synthesis, or communication could be improved.\\
   \hspace{.2in}- Actively scan for logical inconsistencies, incorrect assumptions, or misaligned interpretations -- even when the answer appears complete. When possible, propose alternative reasoning paths or reframe ambiguous user intent to surface potential misunderstandings.\\
   \hspace{.2in}- Your suggestions for the next round should focus on improving the quality of the final answer and should closely align with the user's query.\\
   \hspace{.2in}- If you are not 100\% confident in the completeness or correctness of the answer, initiate a next round of reasoning or agent processing.\\\medskip

Task 3. Determine and recommend next steps:\\
   \hspace{.2in}- Always state whether further agent processing is needed.\\
   \hspace{.2in}- You must verify whether the final answer meets the format requirements specified in the user's query.\\
   \hspace{.2in}- In every case, regardless of output quality, provide concrete suggestions for improvement—such as refining agent prompts, re-evaluating multimodal inputs, or clarifying ambiguous reasoning steps.\\
   \hspace{.2in}- Your output must always move the understanding forward, even when the answer is not yet final.\\\medskip

Background:\\
1. This is the {\color{blue}\{cur\_round\_num\}} round of decision.\\
2. Modality of user's input with short summaries:\\
{\color{blue}\{input\_summaries\}}\\
3. Results of agents and decision of previous rounds.\\
{\color{blue}\{historical\_message\}}\\
4. Agent Pool:\\
{\color{blue}\{available\_agent\_info\}}\\\medskip

Guidelines:\\
- Never repeat agent responses verbatim. Always distill and integrate their content into a unified, user-focused answer.\\
- Whether the output is marked as final or not, you must always provide actionable recommendations to improve the analysis or clarity of the answer.\\
- Be strictly faithful to the user's original query and intent.\\
  \hspace{.2in}- Do not speculate, over-extend, or introduce unrelated or unnecessary information.\\
  \hspace{.2in}- Only answer the user's query; do not add context the user didn't ask for.\\
\end{tcolorbox}
\vspace{-.2in}
\caption{The prompt template of decision stage.}
\label{fig:decision-prompt-template}
\vspace{-.1in}
\end{figure*}

\begin{figure*}[t]
\input{code/master_reasoning_code}
\caption{The \texttt{JSON} schema of reasoning stage output.}
\label{fig:master_reasoning_code}
\end{figure*}

\begin{figure*}[t]
\input{code/master_decision_code}
\caption{The \texttt{JSON} schema of decision stage output.}
\label{fig:master_decision_code}
\end{figure*}

\end{document}